\documentclass[conference]{IEEEtran}

\IEEEoverridecommandlockouts
\usepackage{cite}
\usepackage{amsthm}
\usepackage{amsmath,amssymb,amsfonts}
\usepackage{graphicx}
\usepackage{textcomp}
\usepackage{xcolor}
\def\BibTeX{{\rm B\kern-.05em{\sc i\kern-.025em b}\kern-.08em
    T\kern-.1667em\lower.7ex\hbox{E}\kern-.125emX}}
    
\makeatletter
\newcommand{\linebreakand}{%
  \end{@IEEEauthorhalign}
  \hfill\mbox{}\par
  \mbox{}\hfill\begin{@IEEEauthorhalign}
}
\makeatother

\usepackage[linesnumbered,ruled,vlined]{algorithm2e}

\SetKwInput{KwInput}{Input}                
\SetKwInput{KwOutput}{Output}              

\begin{document}
\title{FedBA: Non-IID Federated Learning Framework in UAV Networks
\\

\thanks{The Heilongjiang Provincial Natural Science Foundation of China (Grant No. LH2020F044), the 2019-``Chunhui Plan'' Cooperative Scientific Research Project of the Ministry of Education of China (Grant No. HLJ2019015), and the Fundamental Research Funds for Heilongjiang University of China (Grant No. 2020-KYYWF-1014) have all contributed to the funding of this work.

*Yi Wu is the corresponding author.}
}

\author{\IEEEauthorblockN{Pei Li}
\IEEEauthorblockA{\textit{School of Data Science and Technology} \\
\textit{Heilongjiang University}\\
Harbin, China \\
2212623@s.hlju.edu.cn}
\and
\IEEEauthorblockN{Zhijun Liu}
\IEEEauthorblockA{\textit{School of Data Science and Technology} \\
\textit{Heilongjiang University}\\
Harbin, China \\
2222708@s.hlju.edu.cn}
\and
\IEEEauthorblockN{Luyi Chang}
\IEEEauthorblockA{\textit{School of Data Science and Technology} \\
\textit{Heilongjiang University}\\
Harbin, China \\
2202518@s.hlju.edu.cn}
\linebreakand 
\IEEEauthorblockN{Jialiang Peng}
\IEEEauthorblockA{\textit{School of Data Science and Technology} \\
\textit{Heilongjiang University}\\
Harbin, China \\
jialiangpeng@hlju.edu.cn}
\and
\IEEEauthorblockN{Yi Wu*}
\IEEEauthorblockA{\textit{School of Data Science and Technology} \\
\textit{Heilongjiang University}\\
Harbin, China \\
1995050@hlju.edu.cn}
}

\maketitle

\begin{abstract}
Advances in artificial intelligence technology and the popularity of the Internet of Things (IoT) devices have brought great convenience to people's lives and significantly improved productivity.
As a new type of Internet of Things device, Unmanned Aerial Vehicles (UAVs) have broad development prospects and have become a hot research field.
However, due to privacy concerns and the limited communication resources of UAVs, it is impractical for UAV devices to transmit their raw data via the air link.
Compared with centralized machine learning, Federated Learning (FL) necessitates the exchange of gradient instead of local data among participating clients and servers, effectively protecting user privacy and reducing the communication cost of participating devices, which is especially suitable for UAV networks.
Nevertheless, there are significant differences in the images captured by different types of drones carrying cameras to different areas (i.e., the problem of statistical heterogeneity), which is still challenging for training FL models.
To this end, we propose an aggregation rule based on the distance between local and global models, named $\mathrm{FedBA}$, to alleviate the problem of data heterogeneity in UAV-assisted FL.
Results from experiments demonstrate that, on three real-world data sets (i.e., CIFAR-100, MNIST, and Fashion-MNIST), our proposed approach performs noticeably better than the conventional FL algorithm.
\end{abstract}


\section{Introduction}
Due to the rapid development of network and communication technologies in recent years, drones have shown the great commercial value and are widely used in traffic monitoring, site management, aerial tourism photography and commercial performances \cite{kouhdaragh2020application}.
Especially with the maturity and perfection of 5G technology and the continuous development of 6G technology, the computing speed as well as data transmission rate has been greatly improved \cite{chowdhury20206g}. Unmanned Aerial Vehicle (UAV) is gradually being used as an aerial service device with its excellent mobility, high computing power, and high perception capability \cite{zhang2020federated}. Drones are uniquely suited for remote and dangerous areas (e.g., wildlife photo collection, atmospheric cloud weather prediction) as well as for scenarios that require extreme real-time performance (e.g., traffic management, mountain power maintenance, disaster prediction). In addition, the UAV cluster, as an aerial edge device, can collect data through its sensors and then transmit the data to the central server for processing and modeling \cite{bithas2019survey}. However, this process can cause privacy leakage of users and increase communication costs \cite{gupta2015survey}. 

The two main reasons for the above problems are as follows. First, when users upload their private data to a third-party central server if the third-party central server is attacked by hackers or the third-party central server is untrustworthy, it will cause the risk of user private data leakage. Second, traditional centralized machine learning requires that a large amount of data be uploaded to a third-party server for computing. Due to the limited communication resources of UAVs, they cannot afford high communication costs \cite{lee2017multi}.

\begin{figure}[!t]
	\centering
	\includegraphics[width=1\linewidth]{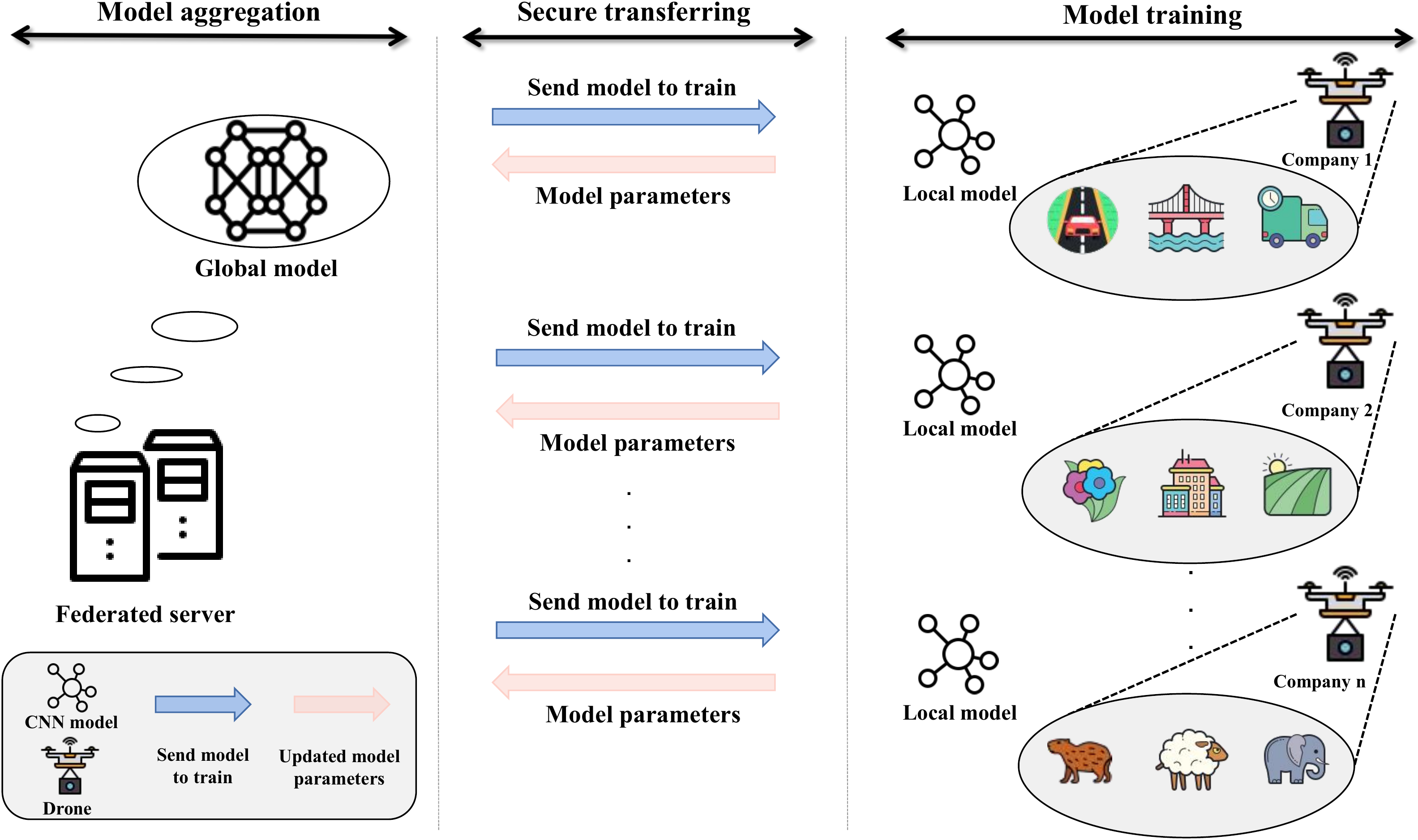}
	\caption{\textcolor{black}{There are three steps in the UAV-assisted federated learning framework approach. First, UAVs from various firms hover at low altitudes to gather pictures of objects in various locations. The UAV employs its computational capability to do local training on the data it has gathered. The drone then runs numerous local model training iterations until the model converges. The UAV then sends the model's parameters for aggregation to a dependable server in the third party to obtain a global model.}}\label{overview}
\end{figure}

In response to these challenges, Federated Learning (FL) \cite{mcmahan2017communication} is gradually becoming the main focus of popular research. By merely uploading model parameters, federated learning addresses the issue of communication costs and lowers the danger of data leaking. Nevertheless, when multiple UAVs work jointly, the difference in the monitoring area of each UAV causes heterogeneity in the collected sensing data. This eventually causes the local device's ability to recognize data and the precision of the global model to decline \cite{sattler2019robust} \cite{zhu2021federated}. 
To this end, we propose a brand-new calculation method for an aggregate based on the distance between training models to palliate the non-IID problem caused by local data.
The system overview of this paper is shown in Fig. \ref{overview}.
Consequently, the following are the contributions to this paper:

\begin{itemize}
\item We propose an FL-enabled framework for the Internet of drones that protects user privacy and reduces communication costs between drones. 
\item We propose a new FL aggregation algorithm named FedBA on non-IID data to cope with the statistical heterogeneity issue in the UAV-assisted FL framework. It is worth noting that this algorithm adds no additional computational cost to the drone.  
\item We assess our algorithm's viability for classification tasks on images from the real world on three datasets (i.e., CIFAR-100, MNIST, and Fashion-MNIST). From the final test results, it is found that our method has better test accuracy than the baseline and has a better advantage in convergence speed.
\end{itemize}

\section{RELATED WORKS}

\subsection{Privacy-preserving UAV Image Recognition}
In the last several years, UAVs have become the main means of acquiring data due to their advantages of multi-view image acquisition and extreme mobility\cite{oleksyn2021going}. For example, using drone image recognition technology for mountainous power grid inspection projects, construction material image recognition, rare animal identification detection, etc. However, in most cases, the images captured by drones are private data. This data cannot be easily transferred to the cloud for centralized training or shared\cite{yao2021secure}. Researchers have applied blockchain and federated learning frameworks to the task of drone image recognition, which allows for the privacy protection of image data captured by drones.
For example, reference \cite{pokhrel2020federated} used blockchain to authorize joint learning of disaster response systems using drones. 
Reference \cite{donevski2021addressing} suggested applying the FedProx \cite{sahu2018convergence} algorithm within the FL framework for usage in driverless traffic monitors to enhance the performance of driverless road transport.
\subsection{Federated Learning on Non-IID Data}
For federated learning, FedAvg [7] is the most representative algorithm. It can be applied to many different real-world scenarios, such as medical image model recognition, financial risk control systems, etc. However, non-IID data still hurts FedAvg. Solving the non-IID problem has also been a topic of great interest in federation learning \cite{li2022federated}. For example, reference\cite{zhao2018federated} created a common public data set using data from local clients. Each client is pre-trained with a pre-prepared public dataset, and then the trained model is trained with its private data to mitigate the impact of data heterogeneity to a certain extent.
Since client data are private, the researchers plan to reduce the discomfort brought on by the non-IID data by creating a unique model for each client as opposed to using the standard global model distributed by the server. 
Reference\cite{li2020federated} adjusted the number of local iterations dynamically and introduces additional regularization terms. In this way, the gap between the server global model and the client local model is continuously narrowed, thereby continuously alleviating data heterogeneity. Reference\cite{arivazhagan2019federated} solved the statistical heterogeneity problem by adopting a personalized federation learning approach, which uses a personalization layer and a base layer in the model to achieve personalization, thus providing a unique model for each client to solve the statistical heterogeneity problem. 
Reference \cite{deng2020adaptive} created personalized models by local fine-tuning to solve non-IID problems.

\section{methodology}
In the present paragraph, we describe the specific formula of federated learning as well as the training process. At the same time, we also describe the specific formula of the algorithm proposed in this article.
\subsection{Federated Learning}
As a distributed learning paradigm, federated learning has gradually entered the public's vision. It can satisfy multiple clients to realize data sharing based on protecting local data security, thereby solving the problem of data silos\cite{chu2021fedfair}. In the entire federated learning process, the transfer of gradients is used to replace the transfer of data, thereby ensuring the security of the data.
At the same time, the client's private data is kept locally throughout the training process and will not be leaked to third-party servers. The goal of our final training is to minimize the global loss $\mathcal{L}(\cdot)$ until it converges, and the following equation is usually used as the objective function:
\begin{equation}\tag{1}
\mathcal{L(\delta)} = \sum\limits_{k=1}^{K}p_{k}\mathcal{L}_{k}(\delta),
\end{equation}
where $\mathcal{L}_k$ represents the $k$-th client's local loss, $p_k$ is the relative contribution rate of the $k$-th client where $\sum\limits_{k=1}^{K}p_{k} = 1$, and $\delta$ stands for model parameters. In the federated average (i.e., FedAvg \cite{mcmahan2017communication}), the global loss function $\mathcal{L(\delta)}$ as follows:
\begin{equation}\tag{2}
\mathcal{L(\delta)}=\sum\limits_{k=1}^{K}{\frac{n_k}{n}}{\mathcal{F}_k}(\delta) \quad where\quad\mathcal{F}_k(\delta)={\frac{1}{n_k}}\sum\limits_{i=1}^{n_k}f_{i}(\delta),
\end{equation}
where $n$ represents all data, $n_k$ is the amount of data owned by the $k$-th client, and $f_{i}(\delta)$ is the cross entropy loss function used to calculate the training loss of the $i$-th sample.

\subsection{Proposed Algorithm}
First, we define a distance function $\mathcal{A}(\|\delta^k - \delta\|^2)$. The purpose of the distance function $\mathcal{A}(\|\delta^k - \delta\|^2)$ is to evaluate the distance between two models, where the global model $\delta$ and the $k$-th local model $\delta^k$.
The function $\mathcal{A}(\cdot)$ has the following characteristics:
\begin{itemize}
\item $\mathcal{A}$ is nonlinear and an increasing convex function when $\|\delta^k-\delta\|^2 \in [0, \infty)$ with $\mathcal{A}(0)=0$.
\item When $\|\delta^k-\delta\|^2 \in [0, \infty)$, $\mathcal{A}$ can be differentiated continuously.
\end{itemize}

Specifically, a typical way to measure the nonlinear function of the distance between $\delta^k$ and $\delta$ is a negative exponential function with both maximum and minimum incentive penalty functions\cite{huang2021personalized}. In this paper, we use the following method to measure the distance between $\delta^k$ and $\delta$.
\begin{equation}\tag{3}
    \mathcal{A}(\|\delta^k-\delta\|^2) = \log[\emph{g}(\|\delta^k-\delta\|^2)].
\end{equation}
Here, \textcolor{black}{we define $g(\cdot) \in [0, \pi/2)$} and the $g(\cdot)$ function formula is as follows:
\begin{equation}\tag{4}
g(x)=\left\{
	\begin{aligned}
	&\emph{x},   &&0\leq x\leq 1\\
	&\arctan(\emph{x}),    &&otherwise
	\end{aligned}
	\right
	.
\end{equation}

Next, we present the FedBA algorithm proposed in this paper.
Suppose there are $K$ clients in the federated learning training. 
The $\mathcal{D}_k$ = $\{{x}_i, {y}_i\}_{i=1}^n$ denotes the local private dataset of client $k$, where ${x}_i$ is the input data sample (e.g., images captured by the UAVs), and ${y}_i$ is the corresponding target output (e.g., recognized category labels). The FedBA algorithm's precise procedure is as follows.
\begin{itemize}
\item \textbf {\textit{Step 1, Local Model Training:}} 
First, in round $t$, each local client receives the aggregated global model from the server side. Second, each local client uses its private dataset to iteratively update the model parameters using Stochastic Gradient Descent (SGD) method. The formula for completing the update for the $k$-th client model is as follows:
\begin{equation}\tag{5}
    \delta^k_{t} \gets \delta_t - \eta\nabla l(k, \delta_t, \mathcal{D}_k,e, b).
\end{equation}
where $\eta$ is the gradient descent algorithm's learning rate, 
$\delta^k_{t}$ is the model after the $k$-th client completes the SGD update in $t$ communication rounds, $\delta_t$ is the global model issued by the server in completing $t$ communication rounds, and $l(\cdot)$ represents the cross entropy loss function.

\item \textbf{\textit{Step 2, Aggregation:}} The server computes the Euclidean distance between each local model and the global model of the previous communication round. Because the range of Euclidean distance values between models spans too large a range and this Euclidean distance function is not a finite function in terms of derivation itself. So we transform the obtained Euclidean distance using the $\mathcal{A}(\cdot)$ function and then use the obtained values to calculate the aggregation weights to obtain the global model for the new round. 
The calculation formula is as follows:
\begin{equation}\tag{6}
    p_{t+1}^{k}=\frac{\mathcal{A}_{k}(\|\delta_{t+1}^{k}-\delta_{t}\|^2)}{\sum\nolimits_{k \in {G_t}} \mathcal{A}_{k}(\|\delta_{t+1}^{k}-\delta_{t}\|^2)},
\end{equation}
where $\mathcal{A}$ is a function of the distance measured between $\delta_{t+1}^{k}$ and $\delta_{t}$, \textcolor{black}{$G_t$ is the collection of round $t$'s participating training clients.}
The server calculates the model parameters received in this communication and obtains the aggregated weight $p_{t+1}^k$ of each client. Afterward, the server aggregates the local models of the client according to the calculated weights to obtain a new global model $\delta_{t+1}$.
The formula is as follows:
\begin{equation}\tag{7}
    \delta_{t+1}= \sum\nolimits_{k \in {G_t}} p_{t+1}^{k} \delta^k_{t+1}.
\end{equation}
\item \textbf{\textit{Step 3, Broadcast Global Model:}} The server broadcasts the next round of global model $\delta_{t+1}$ to each client to continue training.
\end{itemize}

Algorithm 1 describes all the processes in which FedBA continuously iterates between the server and the client.
 \begin{algorithm}
  \caption{FedBA}
  \textbf{Input:} The number of clients is $K$, the sampling rate of the client is $C$, the mini-batch size $\mathcal{B}$ and the number of updates of the local model is $E$, and the learning rate during the SGD of the model is $\eta$.\\
  \textbf{Output:} Training to achieve convergence of the global model $\delta^{*}$.\\
  \textbf{On the server side:}\\
  Initialize a model parameter $\delta_0$\\
  \For{iteration round $t = 1,2,\ldots$}
    {
    $G_t \gets$ $random$ $set$ $of$ $m=max(C\cdot K, 1)$ $clients$\\
    \For{each client k $\in$ $G_t$ in paraller}
      {
      $\delta^{k}_{t+1} \gets \delta^{k}_{t}$\\
$p_{t+1}^{k}=\frac{\mathcal{A}_{k}(\|\delta_{t+1}^{k}-\delta_{t}\|^2)}{\sum\nolimits_{k \in {G_t}} \mathcal{A}_{k}(\|\delta_{t+1}^{k}-\delta_{t}\|^2)}$\\

      $\delta_{t+1}= \sum\nolimits_{k \in {G_t}} p_{t+1}^{k} \delta^k_{t+1}$
      }
    Broadcast $\delta_{t+1}$ to each client
    }
   \textbf{In the $k$-th client:}\\
   \For{local training iteration round $\emph{e} \gets$ $1$ to \emph{E}}
     {
   	\For{mini-batch size b $\in\mathcal{B}$ }
   	  {
   	$\delta^k_{t} \gets \delta_t - \eta\nabla l(k, \delta_t, \mathcal{D}_k, e, b)$\\
   	  }
   	Return $\delta^{k}_{t}$ to server
     }
\end{algorithm}

\begin{figure}[!t]
	\centering
        \includegraphics[width=1\linewidth]{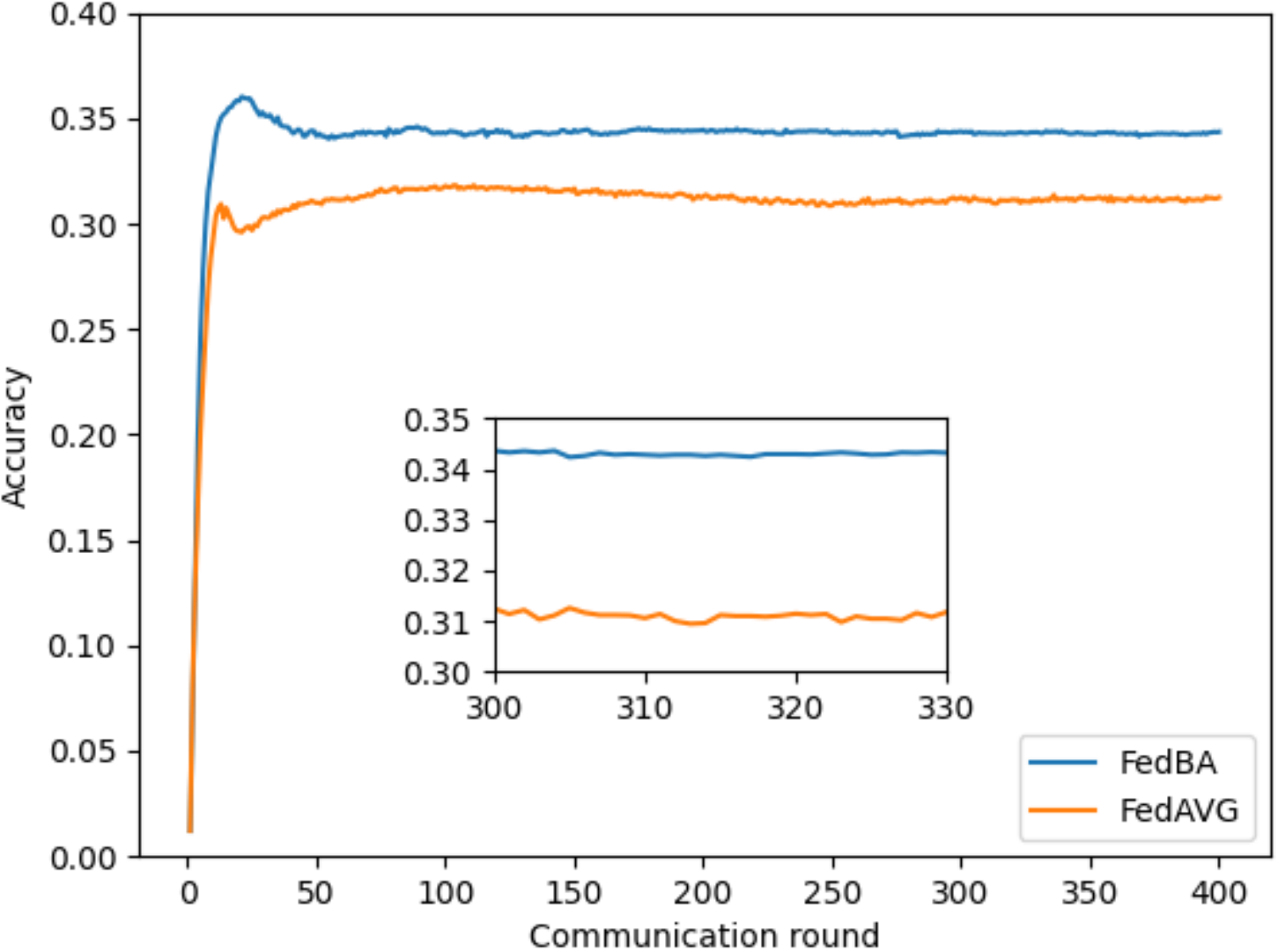}
	\caption{\textcolor{black}{Test accuracy curve on CIFAR-100 dataset.}}\label{ciar100}
\end{figure}

\begin{figure}[!t]
	\centering
        \includegraphics[width=1\linewidth]{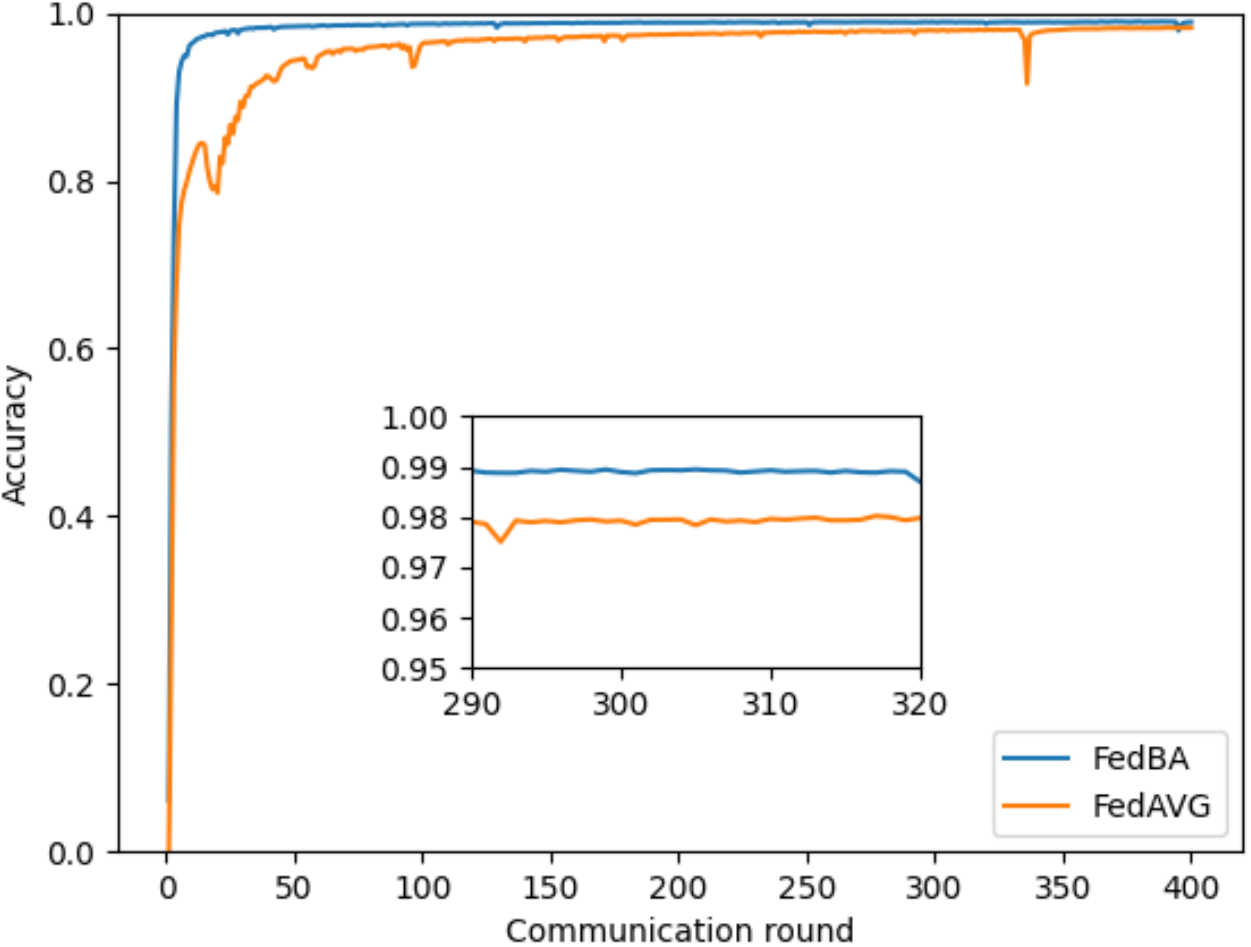}
	\caption{\textcolor{black}{Test accuracy curve on MNIST dataset.}}\label{MNIST}
\end{figure}
\begin{figure}[!t]
	\centering
        \includegraphics[width=1\linewidth]{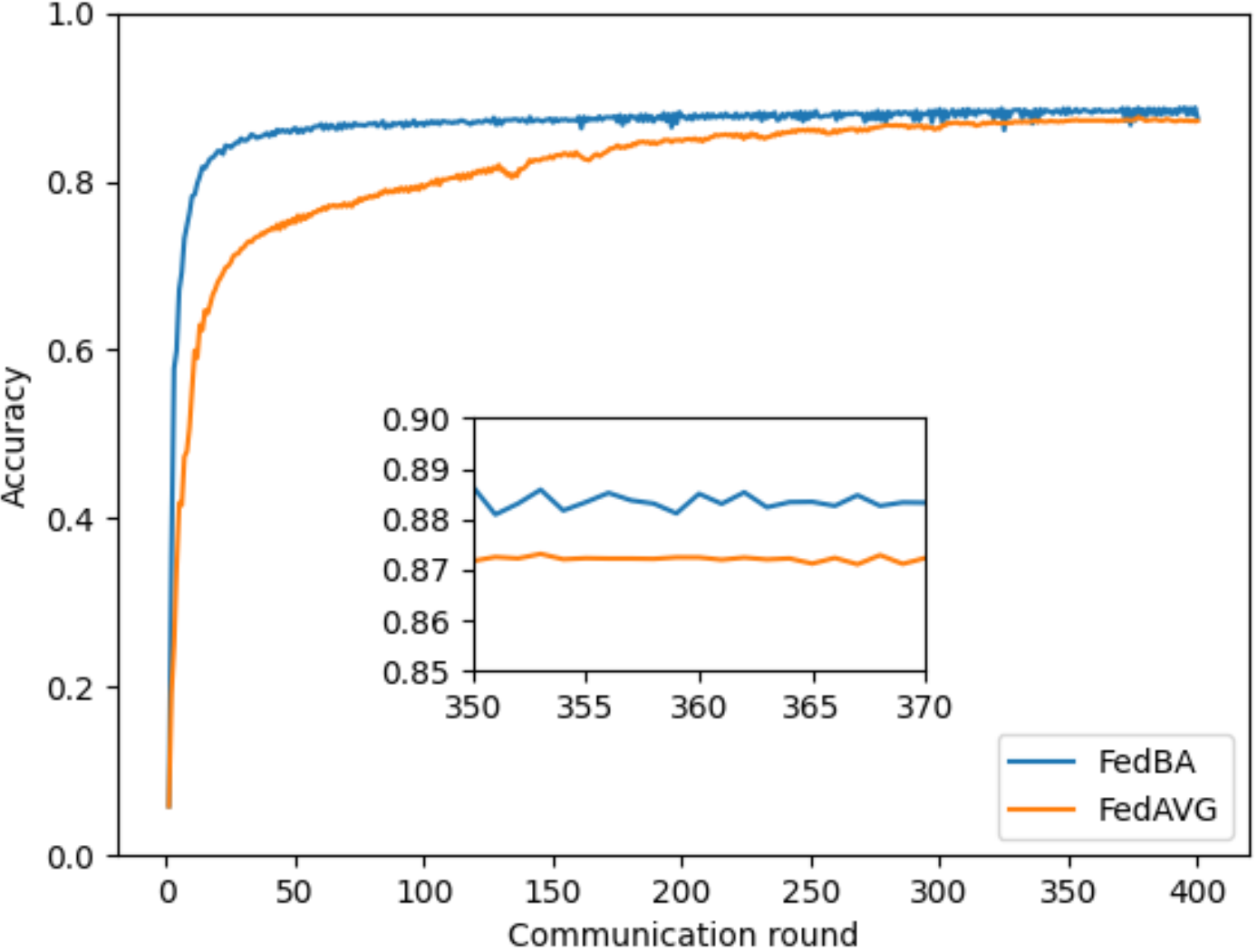}
	\caption{\textcolor{black}{Test accuracy curve on Fashion-MNIST dataset.}}\label{fMNIST}
\end{figure}

\section{EXPERIMENT}
In this subsection, we assess the effectiveness of our proposed framework and contrast it with FedAvg \cite{mcmahan2017communication}. Our experimental environment is configured with Python version 3.9.12 and Pytorch version 1.12.1.
\subsection{Experiment Setup}
We use three real-world datasets CIFAR-100$\footnote{http://www.cs.toronto.edu/~kriz/cifar.html}$, MNIST$\footnote{http://yann.lecun.com/exdb/MNIST}$, and Fashion-MNIST$\footnote{https://github.com/zalandoresearch/Fashion-MNIST}$ to simulate the images collected during drone operations. 
The description of the dataset is as follows: the CIFAR-100 dataset contains 100 common types of animals and objects in life and each category contains 600 pictures and 10,000 pictures are selected as the test set. During the training process, each of our clients was assigned 2500 images as a training set. The MNIST dataset contains a total of 70,000 digit pictures of 10 categories in various handwritten versions and includes 10,000 pictures in the test set. Also during the training process, we allocated 3000 images to each client as a training set. For the Fashion-MNIST dataset, we allocated 3,000 training data to each customer during the entire training process and the overall data set, contains a total of 70,000 data, and 10 categories belong to the category of clothes. At the same time, we adopted the most mainstream non-IID division method, that is, Dirichlet segmentation. We use the Dirichlet distribution function to split the data set, and divide the data set to all local clients at one time. To achieve the simulation of data heterogeneity when UAVs collect pictures in different areas.
By modifying the Dirichlet distribution function's parameter (i.e., $\mu$), it is feasible to influence the degree of non-IID data to mimic the distribution of real-world data. Therefore, we adjust the parameter $\mu$ so that the sample labels are distributed differently on each client.
More precisely, the samples of each category label in our experimental data were distributed among different clients according to different proportions so that each client contained all category labels, but with different amounts of label data.
\subsection{Training Details} In the experimental process of this paper, we use SGD to update a six-layer (convolutional layer, pooling layer, and fully connected layer each occupying 2 layers) Convolutional Neural Network (CNN) model.
In addition, we selected $K=20$ as the maximum number of clients, the sampling rate of clients $C=0.6$, the learning rate in gradient descent $\eta=1e-3$, the number of local epochs $E=5$, the mini-batch size $\mathcal{B} = 64$, and the parameters of the Dirichlet distribution function $\mu=1e-1$.
\subsection{Experiment Result}
We evaluate the FedBA algorithm on the non-IID settings with the CIFAR-100, Fashion-MNIST, and MNIST datasets. The main advantage of our experiment is mainly reflected in the better test accuracy achieved in the experiment, and at the same time, the experimental results show that our algorithm has a better convergence speed than the baseline, which is more important in the field of UAVs. The analysis of accuracy is detailed below.
\begin{itemize}
    \item As shown in Fig. \ref{ciar100}, we evaluated the accuracy of FedBA in the CIFAR-100 dataset non-IID scenario. With the increasing number of communication rounds, the performance of our model continues to improve and the accuracy rate reaches 34.29\%, and the baseline final accuracy only converges to 31.07\%. Compared to the baseline FedAvg, the accuracy of FedBA was improved by 3.22\%. This is caused by the fact that the baseline FedAvg uses only the simple idea of average aggregation, which does not take into account the different weights assigned to each client in the process of average aggregation.
    \item Fig. \ref{MNIST} shows the accuracy of FedBA in the non-IID scenario of the MNIST dataset. When the final training model converges, we can see that FedBA improves by 0.85\% compared to FedAvg. The result made a small contribution to the improvement, this is because the MNIST dataset is much simpler and the accuracy is very high. It is difficult to make a large improvement.
    \item As shown in Fig. \ref{fMNIST}, we evaluated the accuracy of FedBA in the Fashion-MNIST dataset non-IID scenario. The accuracy of our algorithm FedBA is higher than the baseline FedAvg and has a faster convergence rate throughout the 500 communication rounds. Specifically, FedBA achieves 88.86\% accuracy at final convergence, yet FedAvg only achieves 87.17\%.
    
\end{itemize}

\section{Conclusion}
In this paper, we propose an FL-enabled framework for the Internet of drones that protects user privacy and reduces communication costs between drones.
At the same time, we modify the aggregation method in federated learning to obtain better accuracy and convergence speed from test experiments. In the improvement of the whole method, we have taken into account the non-independent and identical distribution problem caused by the UAV collecting data from different areas. Therefore, it is reasonable to assume that different local UAV models should be assigned different aggregation weights in the federated learning aggregation phase to obtain a higher-quality global model. Overall, our method has achieved good experimental results, but we can continue to conduct further research in the direction of model encryption in the future.

\bibliography{ref}

 
\end{document}